\documentclass[english,communication letter]{IEEEtran}
\usepackage[T1]{fontenc}
\usepackage[latin9]{inputenc}
\usepackage{array}
\usepackage{amsmath}
\usepackage{amsthm}
\usepackage{amssymb}
\usepackage{stackrel}
\usepackage{graphicx}

\makeatletter

\providecommand{\tabularnewline}{\\}

\theoremstyle{plain}
\newtheorem{thm}{\protect\theoremname}
\theoremstyle{remark}
\newtheorem{rem}[thm]{\protect\remarkname}

\usepackage[caption=false, font=footnotesize]{subfig}

\usepackage{algorithmic}
\usepackage{cite}

\ifdefined\showcaptionsetup
 \PassOptionsToPackage{caption=false}{subfig}
\fi
\usepackage{subfig}
\makeatother

\usepackage{babel}
\providecommand{\remarkname}{Remark}
\providecommand{\theoremname}{Theorem}

\begin{document}
\title{A Lightweight RL-Driven Deep Unfolding Network for Robust WMMSE Precoding
in Massive MU-MIMO-OFDM Systems}
\author{{\normalsize Kexuan Wang and An Liu, }{\normalsize\textit{Senior Member,
IEEE}}{\normalsize\thanks{Kexuan Wang and An Liu are with the College of Information Science
and Electronic Engineering, Zhejiang University, Hangzhou 310027,
China (email: \{kexuanWang, anliu\}@zju.edu.cn). \textit{(Corresponding
Author: An Liu)}}}\vspace{-0.7cm}
}
\maketitle
\begin{abstract}
Weighted Minimum Mean Square Error (WMMSE) precoding is widely recognized
for its near-optimal weighted sum rate performance. However, its practical
deployment in massive multi-user (MU) multiple-input multiple-output
(MIMO) orthogonal frequency-division multiplexing (OFDM) systems is
hindered by the assumption of perfect channel state information (CSI)
and high computational complexity. To address these issues, we first
develop a wideband stochastic WMMSE (SWMMSE) algorithm that iteratively
maximizes the ergodic weighted sum-rate (EWSR) under imperfect CSI.
Building on this, we propose a lightweight reinforcement learning
(RL)-driven deep unfolding (DU) network (RLDDU-Net), where each SWMMSE
iteration is mapped to a network layer. Specifically, its DU module
integrates approximation techniques and leverages beam-domain sparsity
as well as frequency-domain subcarrier correlation, significantly
accelerating convergence and reducing computational overhead. Furthermore,
the RL module adaptively adjusts the network depth and generates compensation
matrices to mitigate approximation errors. Simulation results under
imperfect CSI demonstrate that RLDDU-Net outperforms existing baselines
in EWSR performance while offering superior computational and convergence
efficiency.
\end{abstract}

\begin{IEEEkeywords}
WMMSE, imperfect CSI, deep-unfolding, DRL. \vspace{-0.5cm}
\end{IEEEkeywords}

\section{Introduction}

The 6G wireless communication adopts massive multi-user (MU) multiple-input
multiple-output (MIMO) and orthogonal frequency-division multiplexing
(OFDM) technologies to support rapid data growth. Weighted Minimum
Mean Square Error (WMMSE) precoding has been widely recognized for
enabling near-optimal weighted sum-rate performance \cite{WMMSE}.
However, the practical deployment of WMMSE in massive MU-MIMO-OFDM
faces two critical challenges \cite{POWMMSE}. First, it heavily relies
on accurate channel state information (CSI), which is often unavailable
due to limited pilot resources and channel aging. Second, its high
computational complexity must be reduced to meet latency and deployment
constraints.

To address the first challenge, \cite{SWMMSE,SWMMSE2,AnAnLu2} modeled
imperfect CSI using Gaussian-distributed posterior channel models
and formulated the robust precoding problem as an ergodic weighted
sum-rate (EWSR) maximization problem under a sum power constraint
(SPC). Specifically, \cite{SWMMSE} and \cite{SWMMSE2} developed
stochastic WMMSE (SWMMSE) based on the sample average approximation
(SAA) method, which alternately optimizes the receiver, weighting,
and precoding matrices using virtual CSI samples drawn from channel
error statistics. \cite{AnAnLu2} iteratively computed deterministic
equivalents of the EWSR and optimized the precoders via the majorization-minimization
(MM) method. However, these methods either incur high per-iteration
complexity scaling cubically with the number of transmit antennas
or suffer from slow convergence due to nested iterations. To address
the second challenge, some robust linear precoding schemes, such as
regularized zero-forcing (RZF) \cite{robust-RZF} and signal-to-leakage-and-noise
ratio (SLNR) precoding \cite{robust-SLNR}, have been proposed but
generally deliver sub-optimal performance compared to WMMSE-based
methods because they do not directly optimize the EWSR objective \cite{AnAnLu2}.
Data-driven approaches such as \cite{HQY12} and \cite{HQY13} employed
multi-layer perceptrons (MLPs) and convolutional neural networks (CNNs)
to approximate the WMMSE algorithm, but suffer from limited interpretability
and generalization. More recently, \cite{POWMMSE} proposed a practical
WMMSE network (PO-WMMSE Net) that combines deep unfolding and approximation
techniques to reduce complexity and accelerate convergence. By leveraging
domain knowledge and introducing only a small number of trainable
compensation matrices, it achieves stable performance across diverse
scenarios. Nevertheless, it employs fixed compensation matrices and
a uniform network depth for all channel conditions, limiting its flexibility
under highly dynamic CSI. Moreover, all the aforementioned methods
are designed for narrowband systems. When extended to wideband systems,
they must be applied independently to each subcarrier, resulting in
prohibitive computational overhead.

To address these challenges, this paper first develops a wideband
variant of the SWMMSE algorithm \cite{SWMMSE} and then enhances it
into a practical-oriented, lightweight reinforcement learning-driven
deep-unfolding network (RLDDU-Net). Specifically, RLDDU-Net consists
of a deep-unfolding (DU) module and a reinforcement learning (RL)
module. The DU module maps each iteration of the wideband robust WMMSE
into a network layer and applies approximation techniques to significantly
reduce computational complexity by leveraging beam-domain sparsity
and subcarrier correlation. These approximations also enable the direct
use of CSI statistics, avoiding the sample average approximation (SAA)
and nested iterations required in conventional robust WMMSE methods,
thereby accelerating convergence. The RL module adaptively generates
compensation matrices to mitigate approximation errors and adjusts
the network depth in response to varying CSI conditions. Numerical
results under imperfect CSI demonstrate that the proposed RLDDU-Net
significantly outperforms baseline approaches.

\section{System Model and Problem Formulation}

\begin{figure}
\centering{}\includegraphics[height=1.9cm]{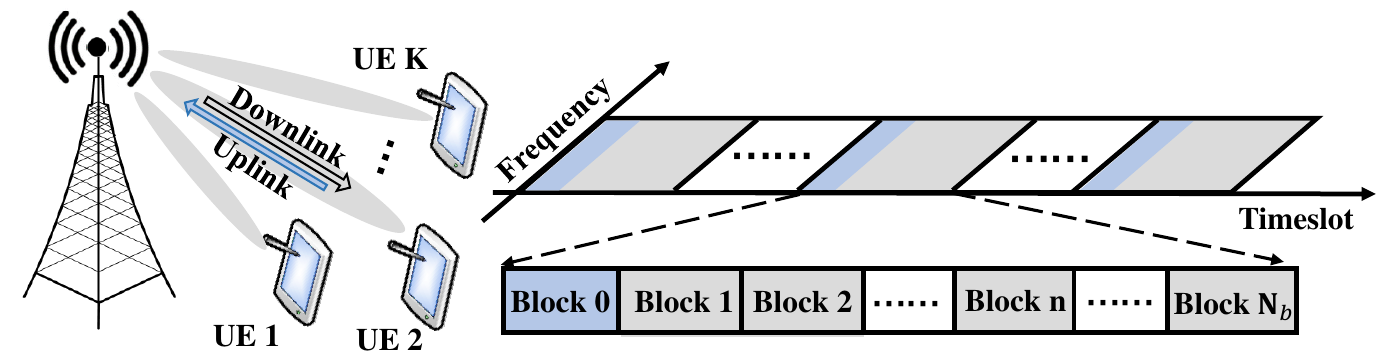}\vspace{-0.3cm}
\caption{\label{fig:System-model}System model and time slot diagram}
\vspace{-0.6cm}
\end{figure}

As illustrated in Fig. \ref{fig:System-model}, we consider a massive
MU-MIMO-OFDM system operating in time-division duplexing (TDD) mode,
where a base station (BS) with a uniform planar array (UPA) of $M_{t}$-antennas
serves $K$ users, each with $M_{r}$-antennas ($M_{r}\ll M_{t}$).
Specifically, the time domain is divided into multiple timeslots,
each further split into $(N_{b}+1)$ blocks: the $0$-th block is
used for uplink training, and the remaining $N_{b}$ blocks are reserved
for downlink transmission. In the frequency domain, subcarriers are
grouped into resource blocks (RBs), each consisting of 12 consecutive
subcarriers. In 5G systems, adjacent $N_{RB}$ RBs (typically $N_{RB}=4$)
are grouped into a resource block group (RBG), which serves as the
granularity for precoding to reduce computational overhead \cite{wideband1}.
We assume block flat fading per subcarrier, i.e., the channel remains
constant within each block. In the following, we consider one timeslot
and one RBG as an example. 

During the uplink training phase (i.e., the $0$-th block), advanced
Bayesian estimation methods exploit beam-domain sparsity to obtain
posterior channel estimates with low pilot overhead \cite{LiXiang_Lian}.
The posterior beam-domain channel at subcarrier $f$ is modeled as
$\mathbf{H}_{k,f,0}^{\mathbf{b}}\sim\mathcal{CN}\bigl(\bar{\mathbf{H}}_{k,f,0}^{\mathbf{b}},\Delta\mathbf{H}_{k,f,0}^{\mathbf{b}}\bigr)$,
where $\bar{\mathbf{H}}_{k,f,0}^{\mathbf{b}}\in\mathbb{C}^{M_{r}\times M_{t}}$
and $\Delta\mathbf{H}_{k,f,0}^{\mathbf{b}}\in\mathbb{C}^{M_{r}\times M_{t}}$
are the posterior mean and covariance. For any downlink block $n$
($n=1,\ldots,N_{b}$), the estimated beam-domain channel at subcarrier
$f$ is denoted as $\mathbf{H}_{k,f,n}^{\mathbf{b}}\in\mathbb{C}^{M_{r}\times M_{t}}$.
To capture channel aging, channel evolution across blocks is modeled
using a first-order Gauss-Markov process \cite{AnAnLu2} as:
\begin{align*}
\mathbf{H}_{k,f,n}^{\mathbf{b}}= & \beta_{k,f,n}\mathbf{\bar{H}}_{k,f,0}^{\mathbf{b}}+\sqrt{1-\beta_{k,f,n}^{2}}\mathbf{M}_{k,f}\odot\mathbf{W}_{k,f,n},\forall k,n,
\end{align*}
where $\odot$ denotes the element-wise multiplication, $\mathbf{M}_{k,f}\in\mathbb{C}^{M_{r}\times M_{t}}$
is a non-negative matrix representing the amplitude profile, and $\mathbf{W}_{k,f,n}\in\mathbb{C}^{M_{r}\times M_{t}}$
is an i.i.d complex Gaussian noise matrix with zero mean and unit
variance entries. The value of $\beta_{k,f,n}\in\bigl(0,1\bigr)$
and $\mathbf{\Omega}_{k,f}\triangleq\mathbf{M}_{k,f}\odot\mathbf{M}_{k,f}$
can be acquired through the sounding process at the BS. Accordingly,
the distribution of $\mathbf{H}_{k,f,n}^{\mathbf{b}}$ is derived
as
\begin{equation}
\mathbf{H}_{k,f,n}^{\mathbf{b}}\sim\mathcal{CN}\bigl(\bar{\mathbf{H}}_{k,f,n}^{\mathbf{b}},\Delta\mathbf{H}_{k,f,n}^{\mathbf{b}}\bigr)\label{eq:distribution H}
\end{equation}
where $\Delta\mathbf{H}_{k,f,n}^{\mathbf{b}}=\beta_{k,f,n}^{2}\Delta\mathbf{H}_{k,f}^{\mathbf{b}}+\bigl(1-\beta_{k,f,n}^{2}\bigr)\mathbf{\Omega}_{k,f}$,
and $\bar{\mathbf{H}}_{k,f,n}^{\mathbf{b}}=\beta_{k,f,n}\mathbf{\bar{H}}_{k,f}^{\mathbf{b}}$.
Without loss of generality, this paper assumes that $\bigl\{\Delta\mathbf{H}_{k,f,n}^{\mathbf{b}},\bar{\mathbf{H}}_{k,f,n}^{\mathbf{b}},\forall k,f,n\bigr\}$
are known and uses them to design the robust precoders.

In the downlink transmission block $n$ ($n\geq1$), the BS applies
an antenna-domain precoder $\mathbf{V}_{k,n}\in\mathbb{C}^{M_{t}\times M_{r}}$
for each user $k$ to process the transmit signal $\boldsymbol{s}_{k,f,n}\in\mathbb{C}^{M_{r}\times1}$,
where $\mathbb{E}\bigl[\boldsymbol{s}_{k,f,n}\boldsymbol{s}_{k,f,n}^{H}\bigr]=\mathbf{I}$.
For simplicity, the block index $n$ is omitted in the following when
no ambiguity arises. The antenna-domain channel at subcarrier $f$
is expressed as $\mathbf{H}_{k,f}=\mathbf{H}_{k,f}^{\mathbf{b}}\mathbf{\Phi}$,
where $\mathbf{\Phi}\in\mathbb{C}^{M_{t}\times M_{t}}$ is the array
response matrix, modeled as a normalized discrete Fourier transform
matrix that satisfies $\mathbf{\Phi}\mathbf{\Phi}^{H}=\mathbf{I}$
for large-scale uniform planar arrays. The received signal at user
$k$ and subcarrier $f$ is\vspace{-0.2cm}
\[
\boldsymbol{y}_{k,f}=\mathbf{H}_{k,f}\mathbf{V}_{k}\boldsymbol{s}_{k,f}+\sum_{m\in\mathcal{K}/\left\{ k\right\} }\mathbf{H}_{k,f}\mathbf{V}_{m}\boldsymbol{s}_{m,f}+\boldsymbol{n}_{k,f},
\]
where $\boldsymbol{n}_{k,f}\sim\mathcal{CN}\bigl(\boldsymbol{0},\sigma_{k}^{2}\mathbf{I}\bigr)$
denotes additive white Gaussian noise, and $\mathcal{K}\triangleq\bigl\{1,2,\ldots,K\bigr\}$
is the set of all user indices. The achievable transmission rate of
user $k$ at subcarrier $f$ is{\small
\[
\mathcal{R}_{k,f}\bigl(\mathbf{V}_{k}\bigr)=\mathrm{log\,det}\biggl(\mathbf{I}+\frac{\mathbf{H}_{k,f}\mathbf{V}_{k}\mathbf{V}_{k}^{H}\mathbf{H}_{k,f}^{H}}{\sum_{m\in\mathcal{K}/\left\{ k\right\} }\mathbf{H}_{k,f}\mathbf{V}_{m}\mathbf{V}_{m}^{H}\mathbf{H}_{k,f}^{H}+\sigma_{k}^{2}\mathbf{I}}\biggr).
\]
}{\small\par}

To enhance system performance under imperfect CSI, the robust precoding
problem is fomulated as the following EWSR maximization problem subject
to a SPC:
\begin{align*}
\mathbf{P}_{1}:\underset{\mathbf{V}}{\mathrm{max}} & \sum_{k\in\mathcal{K}}\sum_{f\in\mathcal{F}}\omega_{k}\mathbb{E}\bigl[\mathcal{R}_{k,f}\bigl(\mathbf{V}_{k}^{i}\bigr)\bigr],\\
\mathrm{s.t.} & \sum_{k\in\mathcal{K}}\mathrm{Tr}\bigl(\mathbf{V}_{k}\mathbf{V}_{k}^{H}\bigr)\leq P_{max}.
\end{align*}
where $\mathcal{F}\triangleq\bigl\{1,2,\ldots,12N\bigr\}$ is the
set of subcarrier indices, $\omega_{k}$ represents the priority of
user $k$, and $\mathbb{E}$ is the expectation taken over distributions
(\ref{eq:distribution H}).

\section{Robust Precoding Algorithm and Network Design}

\subsection{A Wideband Variant of SWMMSE Algorithm \label{subsec:Iterative-Optimization-Based-Alg}}

We first develops a wideband variant of the SWMMSE in \cite{SWMMSE}
to solve $\mathbf{P}_{1}$. To avoid directly handling the constraint,
we incorporate it into the objective and consider problem $\mathbf{P}_{2}$:\vspace{-0.2cm}
\begin{align*}
\mathbf{P}_{2}:\underset{\mathbf{V}}{\mathrm{max}} & \text{\ensuremath{\sum_{k\in\mathcal{K}}\sum_{f\in\mathcal{F}}\omega_{k}}\ensuremath{\mathbb{E}\bigl[\mathcal{\tilde{R}}_{k,f}\bigr]}},
\end{align*}
where\vspace{-0.2cm}
\begin{align}
\mathcal{\widetilde{R}}_{k,f}=\mathrm{log\,det}\Bigl( & \mathbf{I}+\mathbf{H}_{k,f}\mathbf{V}_{k}\mathbf{V}_{k}^{H}\mathbf{H}_{k,f}^{H}\Bigl(\sum_{m\in\mathcal{K}/\left\{ k\right\} }\mathbf{H}_{k,f}\mathbf{V}_{m}\nonumber \\
\times\mathbf{V}_{m}^{H}\mathbf{H}_{k,f}^{H} & +\frac{\sigma_{k}^{2}}{P_{max}}\stackrel[m=1]{K}{\sum}\mathrm{Tr}\bigl(\mathbf{V}_{m}\mathbf{V}_{m}^{H}\bigr)\mathbf{I}\Bigr)^{-1}\Bigr).\label{eq:rate_tilde}
\end{align}
According to \cite{POWMMSE}, any stationary point $\mathbf{V}_{k}^{**}$
of $\mathbf{P}_{1}$ corresponds to a stationary point $\mathbf{V}_{k}^{*}$
of $\mathbf{P}_{2}$, satisfying $\mathbf{V}_{k}^{**}=\xi\mathbf{V}_{k}^{*}$
with $\xi=\sqrt{P_{max}}\bigl(\sum_{k=1}^{K}\mathrm{Tr}\bigl(\mathbf{V}_{k}^{*}\bigl(\mathbf{V}_{k}^{*}\bigr)^{H}\bigr)\bigr){}^{-\frac{1}{2}}$.
Therefore, $\mathbf{P}_{1}$ can be solved by addressing $\mathbf{P}_{2}$,
followed by a scaling operation.

Then, according to \cite[Lemma 1]{SWMMSE}, $\mathbf{P}_{2}$ can
be equivalently reformulated as the following weighted MMSE problem
$\mathbf{P}_{3}$ by introducing auxiliary variables $\bigl\{\mathbf{U}_{k,f},\mathbf{W}_{k,f}\bigr\}$:\vspace{-0.2cm}

{\small
\begin{align*}
\mathbf{P}_{3}:\underset{\mathbf{V}}{\mathrm{min}} & \mathbb{E}\Bigl[\underset{\left\{ \mathbf{U},\mathbf{W}\right\} }{\mathrm{min}}\sum_{k\in\mathcal{K}}\sum_{f\in\mathcal{F}}\omega_{k}\Bigl(\mathrm{Tr}\bigl(\mathbf{W}_{k,f}\mathbf{\tilde{E}}_{k,f}\bigr)-\mathrm{log\,det}\mathbf{W}_{k,f}\Bigr)\Bigr],
\end{align*}
}where the MSE matrix $\mathbf{\tilde{E}}_{k,f}$ is given by\vspace{-0.2cm}
\begin{align*}
\mathbf{\tilde{E}}_{k,f} & \triangleq\bigl(\mathbf{I}-\mathbf{U}_{k,f}^{H}\mathbf{H}_{k,f}\mathbf{V}_{k}\bigr)\bigl(\mathbf{I}-\mathbf{U}_{k,f}^{H}\mathbf{H}_{k,f}\mathbf{V}_{k}\bigr)^{H}\\
+ & \sum_{m\in\mathcal{K}/\left\{ k\right\} }\mathbf{U}_{k,f}^{H}\mathbf{H}_{k,f}\mathbf{V}_{m}\mathbf{V}_{m}^{H}\mathbf{H}_{k,f}^{H}\mathbf{U}_{k,f}+\sigma_{k}^{2}\mathbf{U}_{k,f}^{H}\mathbf{U}_{k,f}.
\end{align*}
Noting that fixing any two of $\mathbf{U},\mathbf{W}$, and $\mathbf{V}$
transforms $\mathbf{P}_{3}$ into a convex problem with a closed-form
solution, we solve $\mathbf{P}_{3}$ using the block coordinate descent
method \cite{WMMSE}. Specifically, at the $i$-th iteration, given
$\mathbf{V}^{i-1}$, the optimal $\mathbf{U}^{i}$ is
\begin{equation}
\mathbf{U}_{k,f}^{i}=\bigl(\mathbf{A}_{k,f}^{i}\bigr)^{-1}\mathbf{H}_{k,f}\mathbf{V}_{k}^{i-1},\forall k,f,\label{eq:U}
\end{equation}
where the variable $\mathbf{A}_{k,f}^{i}=\sum_{m=1}^{K}\mathbf{H}_{k,f}\mathbf{V}_{m}^{i-1}\bigl(\mathbf{V}_{m}^{i-1}\bigr)^{H}\mathbf{H}_{k,f}^{H}+\frac{\sigma_{k}^{2}}{P_{max}}\sum_{m=1}^{K}\mathrm{Tr}\bigl(\mathbf{V}_{m}^{i-1}\bigl(\mathbf{V}_{m}^{i-1}\bigr)^{H}\bigr)\mathbf{I}.$
Once $\mathbf{U}^{i}$ is further fixed, the optimal weight matrix
$\mathbf{W}^{i}$ is
\begin{equation}
\mathbf{W}_{k,f}^{i}=\bigl(\mathbf{I}-\bigl(\mathbf{U}_{k,f}^{i}\bigr)^{H}\mathbf{H}_{k,f}\mathbf{V}_{k}\bigr)^{-1},\forall k,f.\label{eq:W}
\end{equation}
Finally, setting the derivative of the objective function of $\mathbf{P}_{2}$
with respect to $\mathbf{V}_{k}$ to zero yields the optimal precoder{\small
\begin{align}
\mathbf{V}_{k}^{i}= & \bigl(\mathbf{B}^{i}\bigr)^{-1}\sum_{f\in\mathcal{F}}\mathbb{E}\bigl[\mathbf{H}_{k,f}^{H}\mathbf{U}_{k,f}^{i}\mathbf{W}_{k,f}^{i}\bigr],\forall k,\label{eq:V}
\end{align}
whe}re $\mathbf{B}^{i}=\sum_{f=1}^{12N}\sum_{m=1}^{K}\mathbb{E}\bigl[\frac{\sigma_{m}^{2}}{P_{max}}\mathrm{Tr}\bigl(\omega_{m}\mathbf{U}_{m,f}^{i}\mathbf{W}_{m,f}^{i}\mathbf{U}_{m,f}^{i\,H}\bigr)\mathbf{I}$
$+\omega_{m}\mathbf{H}_{m,f}^{H}\mathbf{U}_{m,f}^{i}\mathbf{W}_{m,f}^{i}\mathbf{U}_{m,f}^{i\,H}\mathbf{H}_{m,f}\bigr]$.
Finally, we calculate (\ref{eq:U})-(\ref{eq:V}) using the SAA method
in \cite{SWMMSE}, i.e., sample a virtual channel realization from
the distribution (\ref{eq:distribution H}) to estimate the value
of (\ref{eq:U})-(\ref{eq:V}). Following the similar theoretical
analysis in \cite{SWMMSE}, it can be proven that it converges to
a stationary point of $\mathbf{P}_{1}$.

\subsection{Lightweight RL-Driven Deep-Unfolding Network}

\begin{figure}
\begin{centering}
\includegraphics[height=2.8cm]{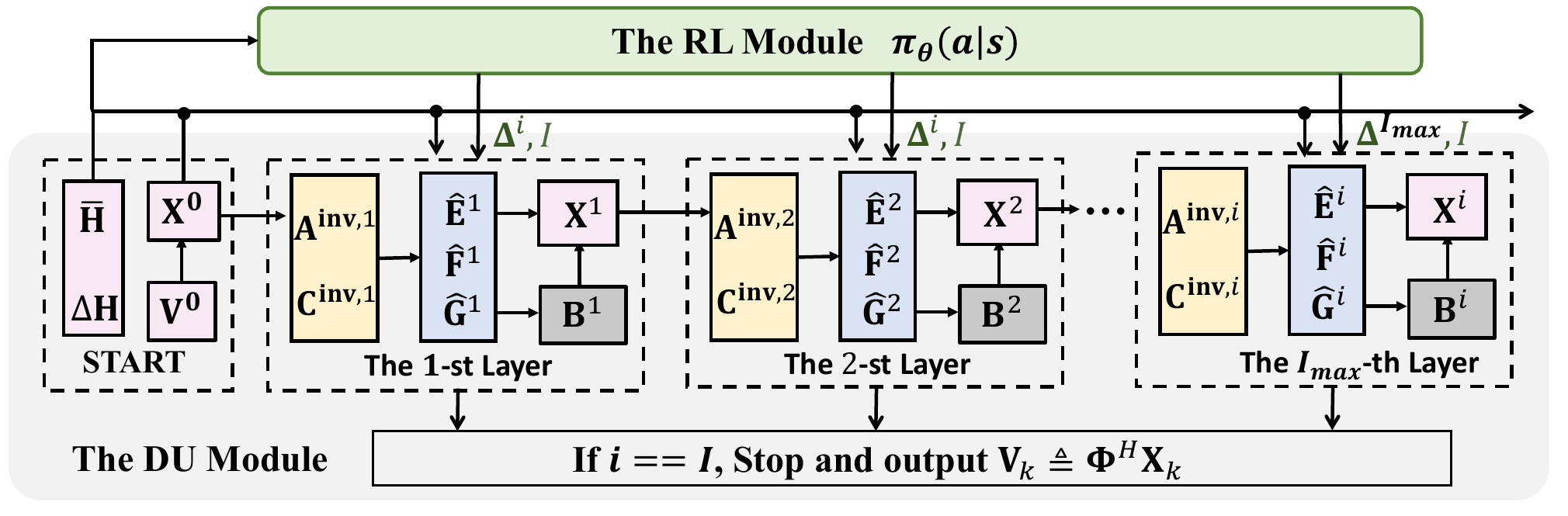}\vspace{-0.3cm}
\caption{\label{fig:network}The RLDDU-Net Architecture.}
\par\end{centering}
\vspace{-0.5cm}
\end{figure}

Building on wideband SWMMSE, we further propose a lightweight RLDDU-Net,
comprising a DU module and a RL module, as shown in Fig. \ref{fig:network}.
The DU module employs approximation techniques for fast convergence
and low complexity, while the RL module uses a policy $\pi_{\boldsymbol{\theta}}$
to adaptively generate compensation matrices and determine the unfolding
depth. We now present the design details. 

\subsubsection{The DU Module \label{subsec:The-Deep-Unfolding-Module}}

In this module, each wideband SWMMSE iteration is first modified and
then mapped into a corresponding network layer. Specifically, to better
exploit beam-domain sparsity, the deep-unfolding module focuses on
optimizing the beam-domain precoders $\bigl\{\mathbf{X}_{k}\triangleq\mathbf{\Phi}\mathbf{V}_{k},\forall k\bigr\}$
instead of $\bigl\{\mathbf{V}_{k},\forall k\bigr\}$. By substituting
(\ref{eq:U}) and (\ref{eq:W}) into (\ref{eq:V}) and applying the
matrix inversion lemma \cite{inverse_lemma}, we obtain that
\begin{equation}
\mathbf{X}_{k}^{i}=\bigl(\tilde{\mathbf{B}}^{i}\bigr){}^{-1}\sum_{f\in\mathcal{F}}\mathbf{E}_{k,f}^{i},\label{eq:X}
\end{equation}
where $\tilde{\mathbf{B}}^{i}=\sum_{f=1}^{N_{f}}\sum_{m=1}^{K}\frac{\sigma_{m}^{2}}{P_{max}}\mathrm{Tr}\bigl(\mathbf{F}_{m,f}^{i}\bigr)\mathbf{I}+\mathbf{G}_{m,f}^{i}$.
By defining $\mathbf{D}_{k,f}^{i}=\mathbf{H}_{k,f}^{\mathbf{b}}\mathbf{X}_{k}^{i-1}\left(\mathbf{X}_{k}^{i-1}\right)^{H}\bigl(\mathbf{H}_{k,f}^{\mathbf{b}}\bigr)^{H}$
and $\mathbf{C}_{k,f}^{i}=\mathbf{A}_{k,f}^{i}-\mathbf{D}_{k,f}^{i}$,
the terms $\bigl\{\mathbf{E}_{k,f}^{i},\mathbf{F}_{k,f}^{i},\mathbf{G}_{k,f}^{i}\bigr\}$
are given by 
\begin{equation}
\mathbf{E}_{k,f}^{i}=\mathbb{E}\bigl[\bigl(\mathbf{H}_{k,f}^{\mathbf{b}}\bigr)^{H}\bigl(\mathbf{C}_{k,f}^{i}\bigr)^{-1}\mathbf{H}_{k,f}^{\mathbf{b}}\bigr]\mathbf{X}_{k}^{i-1},\label{eq:E}
\end{equation}
\begin{equation}
\mathbf{F}_{k,f}^{i}=\mathbb{E}\bigl[\bigl(\mathbf{C}_{k,f}^{i}\bigr)^{-1}\mathbf{D}_{k,f}^{i}\bigl(\mathbf{A}_{k,f}^{i}\bigr)^{-1}\bigr],\label{eq:F}
\end{equation}
\begin{equation}
\mathbf{G}_{k,f}^{i}=\mathbb{E}\bigl[\bigl(\mathbf{H}_{k,f}^{\mathbf{b}}\bigr)^{H}\bigl(\mathbf{C}_{k,f}^{i}\bigr)^{-1}\mathbf{D}_{k,f}^{i}\bigl(\mathbf{A}_{k,f}^{i}\bigr)^{-1}\mathbf{H}_{k,f}^{\mathbf{b}}\bigr].\label{eq:G}
\end{equation}
Note that the precoders $\bigl\{\mathbf{X}_{k}^{i},\forall k\bigr\}$
are fully determined by the three expectation terms $\bigl\{\mathbf{E}_{k,f}^{i},\mathbf{F}_{k,f}^{i},\mathbf{G}_{k,f}^{i}\bigr\}$,
which, however, lack closed-form expressions. Although these terms
can also be estimated via the SAA method, its convergence is slow
and requires numerous channel realizations under high uncertainty
(e.g., large $\Delta\mathbf{H}_{k,f,n}^{\mathbf{b}}$), resulting
in high computational cost. To improve efficiency, we derive closed-form
approximations for $\bigl\{\mathbf{E}_{k,f}^{i},\mathbf{F}_{k,f}^{i},\mathbf{G}_{k,f}^{i}\bigr\}$
based on the statistical information $\bar{\mathbf{H}}_{k,f,n}^{\mathbf{b}}$
and $\Delta\mathbf{H}_{k,f,n}^{\mathbf{b}}$, as detailed below.

Considering that $\mathbf{A}_{k,f}^{i}$ and $\mathbf{C}_{k,f}^{i}$
tend to be diagonal when $M_{t}\gg M_{r}$, we approximate $\mathbb{E}\bigl[\bigl(\mathbf{A}_{k,f}^{i}\bigr)^{-1}\bigl]$
and $\mathbb{E}\bigl[\bigl(\mathbf{C}_{k,f}^{i}\bigr)^{-1}\bigl]$
by by $\mathbf{A}_{k,f}^{\mathbf{inv},i}+\mathbf{Z}_{k,f}^{\mathbf{A},i}$
and $\mathbf{C}_{k,f}^{\mathbf{inv},i}+\mathbf{Z}_{k,f}^{\mathbf{C},i}$,
respectively. Here, $\mathbf{A}_{k,f}^{\mathbf{inv},i}$ and $\mathbf{C}_{k,f}^{\mathbf{inv},i}$
are first-order diagonal Taylor approximations for $\mathbb{E}\bigl[\bigl(\mathbf{A}_{k,f}^{i}\bigr)^{-1}\bigl]$
and $\mathbb{E}\bigl[\bigl(\mathbf{C}_{k,f}^{i}\bigr)^{-1}\bigl]$,
given by:
\begin{equation}
\mathbf{A}_{k,f}^{\mathbf{inv},i}=2\mathbb{E}\bigl[\mathbf{A}_{k,f}^{i}\bigl]^{+}-\mathbb{E}\bigl[\mathbf{A}_{k,f}^{i}\bigl]^{+}\mathbb{E}\bigl[\mathbf{A}_{k,f}^{i}\bigl]\mathbb{E}\bigl[\mathbf{A}_{k,f}^{i}\bigl]^{+},\label{eq:inverse-appro-A}
\end{equation}
\begin{equation}
\mathbf{C}_{k,f}^{\mathbf{inv},i}=2\mathbb{E}\bigl[\mathbf{C}_{k,f}^{i}\bigl]^{+}-\mathbb{E}\bigl[\mathbf{C}_{k,f}^{i}\bigl]^{+}\mathbb{E}\bigl[\mathbf{C}_{k,f}^{i}\bigl]\mathbb{E}\bigl[\mathbf{C}_{k,f}^{i}\bigl]^{+},\label{eq:inverse-appro-C}
\end{equation}
where $\mathbf{Y}^{+}$ denotes the element-wise reciprocal of $\mathbf{Y}$'
diagonal entries with off-diagonal elements set to zero, and compensation
matrices $\bigl\{\mathbf{Z}_{k,f}^{\mathbf{A},i},\mathbf{Z}_{k,f}^{\mathbf{C},i}\bigr\}$
are introduced to narrow the gap between $\bigl\{\mathbf{A}_{k,f}^{\mathbf{inv},i},\mathbf{C}_{k,f}^{\mathbf{inv},i}\bigr\}$
and $\bigl\{\mathbb{E}\bigl[\bigl(\mathbf{A}_{k,f}^{i}\bigr)^{-1}\bigl],\mathbb{E}\bigl[\bigl(\mathbf{C}_{k,f}^{i}\bigr)^{-1}\bigl]\bigr\}$.
Then, using compensation matrices $\bigl\{\mathbf{O}_{k,f}^{\mathbf{E},i},\mathbf{O}_{k,f}^{\mathbf{F},i},\mathbf{O}_{k,f}^{\mathbf{E},i}\bigr\}$
to replace trivial cross-related terms, the expectation terms (\ref{eq:E})-(\ref{eq:G})
are approximated as:
\begin{equation}
\mathbf{\hat{E}}_{k,f}^{i}=\mathbb{E}\bigl[\bigl(\mathbf{H}_{k,f}^{\mathbf{b}}\bigr)^{H}\left(\mathbf{C}_{k,f}^{\mathbf{inv},i}+\mathbf{O}_{k,f}^{\mathbf{E},i}\right)\mathbf{H}_{k,f}^{\mathbf{b}}\bigr]\mathbf{X}_{k}^{i-1},\label{eq:E-appro}
\end{equation}
\begin{equation}
\hat{\mathbf{F}}_{k,f}^{i}=\mathbf{C}_{k,f}^{\mathbf{inv},i}\mathbb{E}\bigl[\mathbf{D}_{k,f}^{i}\bigr]\mathbf{A}_{k,f}^{\mathbf{inv},i}+\mathbf{O}_{k,f}^{\mathbf{F},i},\label{eq:F-appro}
\end{equation}
\begin{equation}
\mathbf{\hat{G}}_{k,f}^{i}=\mathbb{E}\bigl[\bigl(\mathbf{H}_{k,f}^{\mathbf{b}}\bigr)^{H}\left(\hat{\mathbf{F}}_{k,f}^{i}+\mathbf{O}_{k,f}^{\mathbf{G},i}\right)\mathbf{H}_{k,f}^{\mathbf{b}}\bigr].\label{eq:G-appro}
\end{equation}
The the compensation matrix set $\boldsymbol{\varDelta}^{i}=\bigl\{\mathbf{Z}_{k,f}^{\mathbf{A},i},\mathbf{Z}_{k,f}^{\mathbf{C},i},\mathbf{O}_{k,f}^{\mathbf{E},i},\mathbf{O}_{k,f}^{\mathbf{F},i},\mathbf{O}_{k,f}^{\mathbf{G},i},\forall k,f\bigr\}$
is generated by the RL module detailed in section \ref{subsec:The-RL-Driven-Module}.
All second-order expectations involved in equations (\ref{eq:inverse-appro-A})-(\ref{eq:G})
admit closed-form expressions, as detailed in our prior work \cite{POWMMSE}.

Note that both $\bar{\mathbf{H}}_{k,f,n}^{\mathbf{b}}$ and $\Delta\mathbf{H}_{k,f,n}^{\mathbf{b}}$
exhibit sparse structures, with dominant energy concentrated in $B\ll M_{t}$
columns. By zeroing negligible entries in $\bar{\mathbf{H}}_{k,f,n}^{\mathbf{b}}$
and $\Delta\mathbf{H}_{k,f,n}^{\mathbf{b}}$, the matrix multiplication
complexity in RLDDU-Net is reduced to $O\bigl(B^{2}\bigr)$, compared
to $O\bigl(M_{t}^{2}\bigr)$ in WMMSE/WR-WMMSE. Similarly, the matrix
$\tilde{\mathbf{B}}^{i}$ has dominant energy along the diagonal and
in only $q<M_{t}$ rows/columns, allowing its inversion to be computed
with complexity $O\bigl(q^{3}+M_{t}\bigr)$, compared to the cost
of $O\bigl(M_{t}^{3}\bigr)$ for computing $\bigl(\mathbf{B}^{i}\bigr)^{-1}$
in WMMSE/WR-WMMSE. Moreover, to further reduce computational cost
by exploiting the strong correlation across adjacent subcarriers,
we uniformly sample a subset $\tilde{\mathcal{F}}\in\mathcal{F}$
and compute $\bigl\{\mathbf{\hat{E}}_{k,f}^{i},\hat{\mathbf{F}}_{k,f}^{i},\mathbf{\hat{G}}_{k,f}^{i}\bigr\}$
only for $f\in\mathcal{F}$. For the remaining $f\in\mathcal{F}\setminus\tilde{\mathcal{F}}$,
we estimate corresponding matrices using a second-order Lagrange interpolation.
For example, given that $\mathbf{\hat{E}}_{k,f_{0}}^{i}$, $\mathbf{\hat{E}}_{k,f_{1}}^{i}$,
and $\mathbf{\hat{E}}_{k,f_{2}}^{i}$ are known at three subcarrier
$f_{0}<f_{1}<f_{2}$, the value of $\mathbf{\hat{E}}_{k,f}^{i}$ at
any intermediate subcarrier $f\in\bigl(f_{0},f_{2}\bigr)$ can be
approximated as:
\begin{equation}
\mathbf{\hat{E}}_{k,f}^{i}=l_{0}\mathbf{\hat{E}}_{k,f_{0}}^{i}+l_{1}\mathbf{\hat{E}}_{k,f_{1}}^{i}+l_{2}\mathbf{\hat{E}}_{k,f_{2}}^{i},
\end{equation}
where the coefficients are defined as $l_{0}=\frac{\left(f-f_{1}\right)\left(f-f_{2}\right)}{\left(f_{0}-f_{1}\right)\left(f_{0}-f_{2}\right)}$,
$l_{1}=\frac{\left(f-f_{0}\right)\left(f-f_{2}\right)}{\left(f_{1}-f_{0}\right)\left(f_{1}-f_{2}\right)}$,
and $l_{2}=\frac{\left(f-f_{0}\right)\left(f-f_{1}\right)}{\left(f_{2}-f_{0}\right)\left(f_{2}-f_{1}\right)}$.
The same interpolation method also applies to $\hat{\mathbf{F}}_{k,f}^{i}$
and $\mathbf{\hat{G}}_{k,f}^{i}$.

\subsubsection{The RL Module \label{subsec:The-RL-Driven-Module}}

The RL module employ a stochastic policy $\pi_{\boldsymbol{\theta}}$
to generate compensation matrices $\boldsymbol{\varDelta}_{m,n}\triangleq\bigl\{\boldsymbol{\varDelta}_{m,n}^{1},\boldsymbol{\varDelta}_{m,n}^{2},\ldots,\boldsymbol{\varDelta}_{m,n}^{I_{max}}\bigr\}$
and stopping coefficients $\boldsymbol{\beta}_{m,n}\triangleq\bigl\{\beta_{m,n}^{1},\beta_{m,n}^{2},\ldots,\beta_{m,n}^{I_{max}}\bigr\}$,
where the DU network depth is set as $I_{m,n}=\mathrm{argmax}_{i}\,\beta_{m,n}^{i}$.
This design keeps all action vector elements continuous, thereby avoiding
the challenges of mixed discrete-continuous optimization. We model
the sequential decision-making process of $\pi_{\boldsymbol{\theta}}$
as a contextual bandit (CB) problem \cite{RLintroduction} and adopt
a model-free RL approach to optimize the policy parameters $\boldsymbol{\theta}$.

Specifically, each downlink block is treated as a single step in the
CB setting. The $n$-th block in the $m$-th timeslot corresponds
to the step $t=mN_{b}+n$. At each step $t$, the environment presents
a context $\boldsymbol{s}_{t}\triangleq\bigl\{\bar{\mathbf{H}}_{k,f,m,n}^{\mathbf{b}},\Delta\mathbf{H}_{k,f,m,n}^{\mathbf{b}},\forall k,f\bigr\}\in\mathcal{S}$,
for which the RL module sample an action $\boldsymbol{a}_{t}$ from
a Gaussian policy $\pi_{\boldsymbol{\theta}}\bigl(\boldsymbol{a}_{t}\mid\boldsymbol{s}_{t}\bigr)$.
The mean and variance the policy distribution are generated by neural
networks $f_{\boldsymbol{\theta}^{\mu}}\bigl(\boldsymbol{s}_{t}\bigr)$
and $f_{\boldsymbol{\theta}^{\delta}}\bigl(\boldsymbol{s}_{t}\bigr)$,
respectively, where $\boldsymbol{\theta}\triangleq\bigl\{\boldsymbol{\theta}^{\mu},\boldsymbol{\theta}^{\delta}\bigr\}$.
In response to the action, the environment returns an immediate reward
defined as $R\left(\boldsymbol{s}_{t},\boldsymbol{a}_{t}\right)=\mathcal{\hat{R}}_{k,f,m,n}-\mathcal{\hat{\bar{R}}}_{k,f,m,n}$,
where $\mathcal{\hat{R}}_{k,f,m,n}$ and $\mathcal{\hat{\bar{R}}}_{k,f,m,n}$
denote the EWSR performance of the $I_{m,n}$-layer RLDDU-Net and
the $I_{max}$-layer DU network without compensation, respectively.
Note that shallow depth $I_{m,n}$ may lead to underfitting due to
insufficient model capacity, whereas an excessively large $I_{m,n}$
can result in overfitting due to the use of too many compensation
matrices. Therefore, the training objective is to optimize the policy
such that the expected long-term reward is maximized: $\underset{\boldsymbol{\theta}}{\mathrm{max}}\frac{1}{T}\sum_{t=0}^{T-1}\text{\ensuremath{\mathbb{E}\bigl[R\left(\boldsymbol{s}_{t},\boldsymbol{a}_{t}\right)\bigr]}}$.

Model-free RL is particularly suitable for solving such a CB problem,
as it directly optimizes the policy using observed rewards without
fitting an explicit reward model. This is especially important here,
since the reward function is deeply nested and highly nonlinear, making
accurate modeling extremely challenging. Specifically, this paper
adopts a stochastic successive convex approximation (SSCA)-based model-free
RL algorithm from our prior work \cite{hybridpolicyLuyuan}, which
constructs and solves convex surrogate problems, and ensures $\pi_{\boldsymbol{\theta}}$
convergence to a stationary point of the CB problem. The training
code is available at: https://github.com/kexuanWang-zju/RLDDU-code.
Although it is theoretically possible to train the policy network
via unsupervised learning by treating EWSR as the loss function, such
methods require first computing the gradient with respect to the action
and then backpropagating through the network to update the policy
parameters. This nested computation is prone to gradient vanishing,
especially in deep architectures. Moreover, in practical systems where
parameters such as channel aging coefficients may drift over time,
RL enables online adaptation, whereas unsupervised learning methods
are typically limited to offline training.
\begin{rem}
Compared with existing robust precoding algorithms, RLDDU-Net offers
three key advantages: 1) RLDDU-Net incorporates domain knowledge from
the wideband SWMMSE algorithm, which is theoretically guaranteed to
converge to a local optimum of problem $\mathbf{P}_{1}$, thereby
achieving more stable performance across diverse scenarios than black-box
networks in \cite{HQY12} and \cite{HQY13}; it employs approximation
techniques that significantly accelerate convergence and reduce the
computational burden of high-dimensional matrix operations and wideband
multi-subcarrier processing, outperforming SAA-based and bi-level
schemes in \cite{SWMMSE,SWMMSE2,AnAnLu2}; and it leverages RL-driven
design to further enhances its representational capacity over conventional
DU schemes with fixed depth and compensation matrices, such as the
PO-WMMSE Net in \cite{POWMMSE}.
\end{rem}

\section{Numerical Results\label{sec:Numerical-Results}}

\begin{figure*}[t]
\centering{}\subfloat[\label{fig:simulation-A}]{\includegraphics[height=3.8cm]{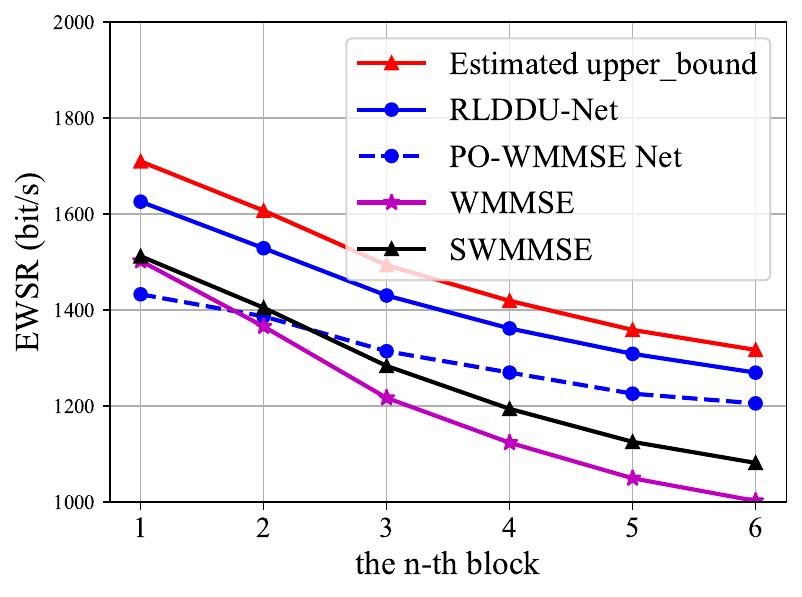} }
\subfloat[\label{fig:simulation-B}]{\includegraphics[height=3.8cm]{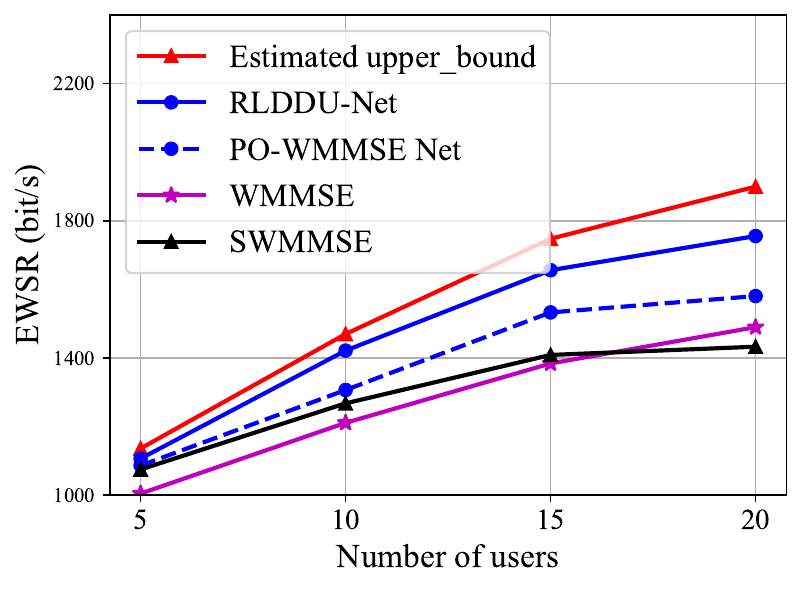}}
\subfloat[\label{fig:simulation-C}]{\includegraphics[height=3.8cm]{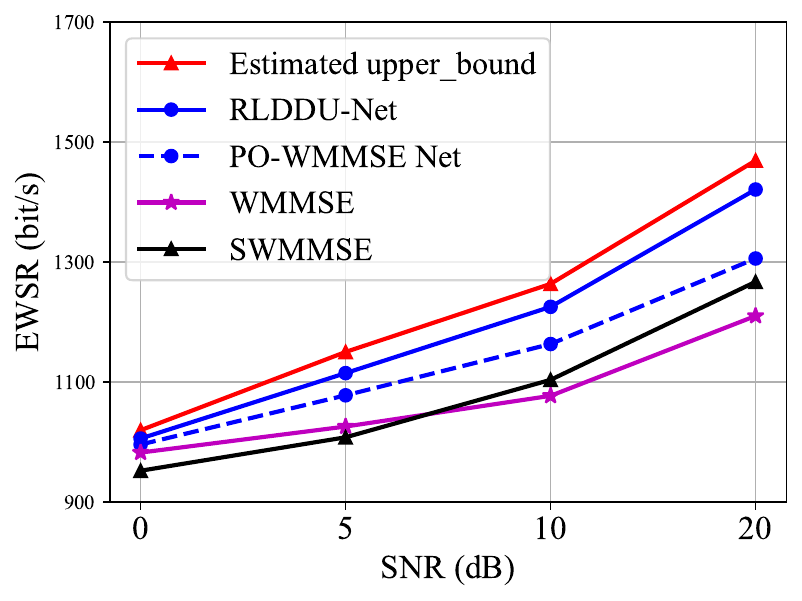}}\vspace{-0.2cm}
\caption{(a) EWSR performance at different block. (b) EWSR performance versus
different number of users. (c) EWSR performance versus different SNRs.}
\vspace{-0.5cm}
\end{figure*}
 We consider a MU-MIMO-OFDM system where a BS with $M_{t}=64$ antennas
serves $K=\bigl\{5,10,15,20\bigr\}$ users, each with $M_{r}=2$ antennas.
Channels are generated using the QuaDRiGa model \cite{AnAnLu5} under
the \textquotedblleft 3GPP\_38.901\_UMa\_NLOS\textquotedblright{}
scenario. Each timeslot lasts 1 ms and is divided in to $N_{b}+1=7$
blocks. The carrier frequency is 3.5\,GHz with a subcarrier spacing
of 15\,kHz. Users are randomly generated within a 200\,m radius around
the BS and move at 120 km/h in random directions. The total transmit
power is $P=1$ W. For simplicity, we assume accurate channel estimation
at the 0-th block, and that CSI inaccuracies are entirely due to channel
aging, characterized by $\alpha$. Following \cite{POWMMSE}, we statistically
obtain $\alpha=\bigl\{0.96,0.92,0.84,0.75,0.63,0.49\bigr\}$ corresponding
to the six downlink blocks, where smaller $\alpha$ indicates more
severe degradation. For RLLDU-Net, we set the the number of sampled
subcarriers frequency-domain interpolation to $\widetilde{F}=8$.
The policy networks $f_{\boldsymbol{\theta}^{\mu}}\bigl(\cdot\bigr)$
and $f_{\boldsymbol{\theta}^{\delta}}\bigl(\cdot\bigr)$ are composed
of two convolutional layers (kernel size $H=3$, channel number $c=8$,
and stride 1) and two fully connected layers (width $d=128$). Baseline
algorithms include wideband the SWMMSE \cite{SWMMSE,SWMMSE2} and
a wideband extension of PO-WMMSE Net \cite{POWMMSE} (i.e, perform
the PO-WMMSE Net on the central subcarrier to obtain the precoder
for the entire RBG), which are referred to as SWMMSE and PO-WMMSE
Net for simplicity in the following. All baselines use 5 iterations/layers,
while the RLDDU-Net adopts a maximum depth of $I_{max}=5$. Additionally,
since wideband SWMMSE can converges to a local optimum of problem
$\mathbf{P}_{1}$ with sufficient iterations, we also simulate it
with 100-iterations to estimate an upper performance bound.

Fig. \ref{fig:simulation-A} illustrates the EWSR performance of algorithms
in a scenario with $K=10$ users and a transmit SNR of 20 dB across
various downlink blocks. The results show that SWMMSE, with sufficient
iterations (i.e., the estimated upper bound), significantly outperforms
WMMSE by accounting for CSI errors. However, with 5 iterations, SWMMSE
provides only a marginal improvement over WMMSE due to slower convergence
from the SAA method. In contrast, RLDDU-Net far exceeds WMMSE\textquoteright s
performance and approaches the upper bound by directly leveraging
CSI statistics (mean and covariance). Simulation statistics reveal
that RLDDU-Net requires an average depth of only $\bigl\{3.2,3.2,3.4,3.7,4.1,4.6\bigr\}$
layers across the six blocks, demonstrating superior convergence efficiency.
Moreover, RLDDU-Net\textquoteright s adaptive compensation matrices
further enhance performance compared to PO-WMMSE Net. While PO-WMMSE
Net, with a fixed compensation matrix, outperforms WMMSE on average,
it performs worse than WMMSE in some blocks. RLDDU-Net consistently
outperforms all other baselines across all blocks, with performance
gains becoming more significant in blocks with less accurate CSI.
Fig. \ref{fig:simulation-B} shows the EWSR performance versus the
number of users. As the number of users increases, the EWSR growth
rate of all algorithms slows down due to higher interference and the
amplified impact of inaccurate CSI. However, the performance gain
of RLDDU-Net over the baselines increases, highlighting its superior
interference management and robustness. Fig. \ref{fig:simulation-C}
shows the performance at different SNRs, where RLDDU-Net consistently
outperforms the baselines and approaches the estimated upper bound.
Moreover, RLDDU-Net shows stronger performance at high SNR compared
to the baselines, as performance at high SNR is primarily limited
by interference, and RLDDU-Net is more effective at mitigating it.

Finally, we compare the computational complexity of RLDDU-Net with
baseline algorithms in Table \ref{tab:Comlexity}, and present numerical
results for the typical setup in Fig. 1. RLDDU-Net exhibits comparable
computational overhead to PO-WMMSE Net, while significantly reducing
complexity compared to conventional WMMSE methods. This is attributed
to two key improvements: 1) Its higher convergence efficiency allows
for fewer layers compared to the number of iterations required by
baseline algorithms, where the average number of iterations of RLDDU-Net
is only $3.7$ in the simulations; 2) By exploiting beam-domain sparsity
(with $B\approx10$ and $q\approx30$, both obtained through simulation
statistics and smaller than $M_{t}=64$) and subcarrier correlation,
the matrix computation cost per layer/iteration is greatly reduced,
resulting in overall complexity scaling linearly with $M_{t}$. 
\begin{table}
\begin{centering}
\caption{Complexity comparison\label{tab:Comlexity}}
\vspace{-0.2cm}
\begin{tabular}{|>{\centering}m{1.5cm}|>{\centering}m{5cm}|>{\centering}m{0.8cm}|}
\hline 
{\scriptsize Algorithms} & {\scriptsize Computational complexity} & {\scriptsize Values}\tabularnewline
\hline 
\hline 
{\scriptsize (S)WMMSE} & {\scriptsize$O\bigl(M_{t}^{2}M_{r}FKI_{max}+M_{t}^{3}I_{max}\bigr)$} & {\scriptsize$2.3e7$}\tabularnewline
\hline 
{\scriptsize PO-WMMSE Net} & {\scriptsize$B^{2}M_{r}^{2}FKI+q^{2}M_{r}KI+\left(M_{t}+q^{3}\right)I$} & {\scriptsize$3e6$}\tabularnewline
\hline 
{\scriptsize RLDDU-Net} & {\scriptsize$B^{2}M_{r}^{2}\tilde{F}KI+q^{2}M_{r}KI+\left(M_{t}+q^{3}\right)I+\left(d+C^{2}\right)H^{2}CBM_{r}\tilde{F}K$} & {\scriptsize$4e6$}\tabularnewline
\hline 
\end{tabular}
\par\end{centering}
\vspace{-0.5cm}
\end{table}

\section{Conclusion}

This paper designs RLDDU-Net for robust WMMSE precoding in massive
MU-MIMO-OFDM systems. Derived from wideband SWMMSE, a robust precoding
algorithm with theoretical guarantees, RLDDU-Net achieves strong performance
under imperfect CSI. By introducing approximation techniques and leveraging
beam-domain sparsity and subcarrier correlation, it greatly reduces
computational complexity and accelerates convergence. Additionally,
the RL-driven network in RLDDU-Net can adaptively generate compensation
matrices to correct approximation errors and adjust network depth.
Simulations demonstrate that RLDDU-Net achieves superior throughput
with significantly reduced computational burden under imperfect CSI,
offering a promising solution for practical WMMSE precoding deployment
in real-world systems.

\bibliographystyle{IEEEtran}
\addcontentsline{toc}{section}{\refname}\bibliography{RLreferences}

\end{document}